\documentclass{article}
\usepackage{spconf,amsmath,graphicx}
\usepackage{algorithm}
\usepackage{algorithmic}
\usepackage{enumerate}

\title{Character Proposal Network FOR ROBUST TEXT EXTRACTION}
\name{Shuye Zhang$^{1}$, Mude Lin$^2$, Tianshui Chen$^2$, Lianwen Jin$^{1,*}$, Liang Lin$^{2}$}
\address{$^{1}$South China University of Technology, China\\
$^{2}$Sun Yat-Sen University, China \\
{\tt $^{*}$lianwen.jin@gmail.com} }

\begin{document}
\ninept
\maketitle
\begin{abstract}

Maximally stable extremal regions (MSER), which is a popular method to generate character proposals/candidates, has shown superior performance in scene text detection. However, the pixel-level operation limits its capability for handling some challenging cases (e.g., multiple connected characters, separated parts of one character and non-uniform illumination). 
To better tackle these cases, we design a character proposal network (CPN) by taking advantage of the high capacity and fast computing of fully convolutional network (FCN).
Specifically, the network simultaneously predicts characterness scores and refines the corresponding locations. The characterness scores can be used for proposal ranking to reject non-character proposals and the refining process aims to obtain the more accurate locations.
Furthermore, considering the situation that different characters have different aspect ratios, we propose a multi-template strategy, designing a refiner for each aspect ratio.  The extensive experiments indicate our method achieves recall rates of 93.88\%, 93.60\% and 96.46\% on ICDAR 2013, SVT and Chinese2k datasets respectively using less than 1000 proposals, demonstrating promising performance of our character proposal network.

\end{abstract}
\section{Introduction}
\label{sec:intro}
Text detection which locates high-level semantic information in natural scene images is of great significance in image understanding and information retrieval. As a part of text detection, character proposal methods which provide a number of candidates that are likely to contain characters, play a key role \cite{neumann2012real}\cite{yin2014robust}\cite{sun2014robust}. Maximally stable extremal regions (MSER) \cite{matas2004robust} which can be viewed as a character proposal method, has achieved superior performance in scene text detection. However, the low-level pixel operation limits its capacity in some challenging cases.
First, if a connected component is comprised of multiple touched characters (e.g., Figure 1(a)), MSER/ER based methods would mistake the proposals containing connected characters as correct proposals \cite{huang2014robust}.
Second, if a single character consists of multiple segmented radicals (e.g., Figure 1(b)), MSER/ER based methods would produce superfluous components, which makes locating character confused. Third, if a character is corrupted by non-uniform illumination, MSER/ER based methods would probably miss it \cite{sun2015robust}.
On the other hand, sliding window based method, which is another method to generate character proposals, is robust to these challenges \cite{jaderberg2014deep}\cite{wang2012end}. However, it costs expensive computation since it requires exhaustive computation on all locations of an image over multiple scales.

\begin{figure}[tb]
\includegraphics[width=0.5\textwidth]{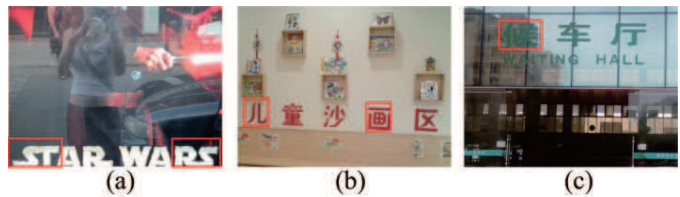}
\caption{The examples corresponding to the three cases: (a) multiple connected characters; (b) separated parts of one character and (c) non-uniform illumination. For clarity, these cases are surrounded by red rectangle boxes.
}
   \label{fig:1}
\end{figure}

In the paper, we design a method for locating character proposals which can make full use of Fully Convolutional Network (FCN). On one hand, FCN is derived from convolutional neural network (CNN) and CNN has shown a superior performance on character and non-character classification \cite{jaderberg2014deep}. On the other hand, it is observed that FCN \cite{long2014fully} takes an image of arbitrary size as input and outputs a dense response map. The receptive field corresponding to each unit of the response map is a patch in original image, hence the intensity of this unit can be treated as a predicted score of corresponding patch.
It approximates a sliding window based method but shares convolutional computation among overlapped image patches, thus it is able to eliminate the redundant computation and make the inference process efficiently.
Based on these analysis, we introduce a novel architecture based on FCN to score the image patches and refine the corresponding locations simultaneously, which we call character proposal network (CPN).
Specifically,
our method can receive a series of input images and output response maps for characterness scores and corresponding locations respectively.
Low-confidence proposals will be filtered by characterness scores and the location response map will output more accurate locations.

Moreover, it is found that different characters have different aspect ratios, so to better tackle the diversity problem of aspect ratios, we propose a multi-template strategy, that is designing a refiner for each aspect-ratio template. Formally, the  characters are first clustered into $K$ templates depending on their aspect ratios. Then, $K$ refiners are designed, while each refiner contains a classifier and a regressor. The classifier outputs a score to suggest whether the patch belongs to the corresponding template and the regressor refines the location based on corresponding aspect ratio. We define a multi-task objective function, including cross-entropy loss and mean square error (MSE) for classifiers and regressors respectively, and optimize CPN in an end-to-end manner.


The remainder of paper is organized as follow. Section 2 introduces the related work. Section 3 describes our character proposal network in details.
Section 4 and 5 illustrate the optimization and inference process respecively. Section 6 presents the experimental results and Section 7 concludes the paper.
\section{Related Work}
\label{sec:format}
Proposal methods for generic object detection have make continuing progress in recent years \cite{uijlings2013selective}\cite{zitnick2014edge}\cite{arbelaez2014multiscale}. However they were not customized for scene text/character.
Character proposal methods have also achieve great improvement. For enhancing the performance of MSER, Koo et al. \cite{koo2013scene} exploited multi-channel techniques (e.g., Y-Cb-Cr channels) and Sun et al. \cite{sun2015robust} proposed a PII color space to enhance the robustness of the MSER/ER based method in non-uniform illumination situation. Neumann and Matas \cite{neumann2012real} proposed to use ER method instead of MSER to improve the performance of MSER/ER based method in blurry and low contrast case. Sun et al. \cite{sun2015robust} proposed contrasting extremal region (CER) to reduce the number of non-text components and repeated text components. Yin et al. \cite{yin2014robust} designed a fast pruning algorithm to extract MSERs as character candidates by using the strategy of minimizing regularized variations. These improvements enhance the performance of MSER/ER based method, however the above mentioned challenges still exist. On the other hand, for promoting the development of sliding window based method, \cite{jaderberg2014deep}\cite{wang2012end} applied convolutional neural network (CNN) as the classifier for distinguishing text and non-text. However, the computational cost is expensive.
Absorbing the benefits of sliding window based method and meanwhile reducing the computation cost, we design a method for locating character proposals. This is clearly different from the work \cite{hosang2015makes}, which applied FCN for semantic segmentation.
Besides, the training of regression network in \cite{sermanet2013overfeat} takes input as the pooled feature maps from layer 5 while our framework is optimized by an end-to-end manner which allows the full back-propagation; the optimization processes of classifier and regressor are implemented separately in \cite{sermanet2013overfeat}, yet in our work they are simultaneously carried out under a unified framework.
\section{Character Proposal Network}
\label{sec:pagestyle}
Given an image patch, character proposal network (CPN) aims to output a score predicting whether the patch containing character and refining locations. Let $S_i$ denote the characterness score of the $i$-th patch $P_i$ and $L_i$ denote corresponding location. Then, our task can be defined as:
\begin{equation}
G(P_i)=\{(S_i, L_i)\},
\end{equation}
where $G(\cdot)$ denotes a convolutional operator. $S_i$ can be used to rank the patches and then select the patches whose confidences are greater than a predefined threshold. $L_i$ can be used to refine the locations of these high-confidence patches.

Simply extracting patches and then feeding them through the network for all patches of a test image, similar to sliding window based method, is computationally very inefficient. Instead, we designed our CPN architecture based on FCN, and it takes a full image (denoted as $I$) as input and output the scores and locations for all patches. For sharing a large amount of convolutional computation among the overlapped patches, CPN is able to compute efficiently and the process can be formulated as:
\begin{equation}
G(I)=\{(S_i, L_i)|_{i=1}^N\}.
\end{equation}

\noindent\textbf{Multiple Template Strategy}
We experimentally found that the aspect ratios of scene characters are varied and CPN tends to regress a biased aspect ratio if using a single aspect ratio under our learning framework. To overcome this problem, a multi-template strategy is proposed. Specifically, different aspect ratios are clustered into $K$ templates. Different classifiers and regressors are customized for these specific templates.

\noindent\textbf{Network Architecture}
Our CPN for English characters is configured as: $29$$\times29$$\times3$ input-96C5S1-P3S2-256C4S1-P3S2-384C3S1-512C2S1-256C1S1-$5K$ output. We refer this network as CPN-ENG. Specifically, an input sample is firstly resized to $29\times29$ size and then fed into the first convolutional layer, which contains 96 convolutional filter kernels of size $5$$\times5$$\times3$ and the stride is 1 pixel (denoted as 96C5S1). The following pooling layer takes 96 previous feature maps as input, and applies 96 max-pooling kernels of size $3\times3$ and the stride is 2 pixels (denoted as P3S2). Likewise, the second, third, fourth and fifth convolutional layer has 256, 384, 512, 256 convolutional kernels respectively. It is worth noting that there is no fully connected layers in CPN-ENG. On the other hand, the CPN architecture for Chinese characters is configured as: $43$$\times43$$\times3$ input-96C5S1-P3S2-256C5S1-P3S2-384C3S1-384C3S1-512C2S1-512C2S1-$5K$ output. We refer this network as CPN-CHS. The size of training image (denoted as $\mathbf{R}=(R_w, R_h)$) in CPN-ENG is larger than that of CPN-CHS, since the Chinese characters with relatively complex structure require a larger receptive field.
We find that the stride of two overlapped receptive fields is a product of the strides of convolutional layer or pooling layer in our network.
Specifically, it is written as $s=\prod_i s_i$
where $s_i$ denotes the stride of convolutional or pooling layer. 
\begin{figure}[tp]
   \includegraphics[width=0.5\textwidth]{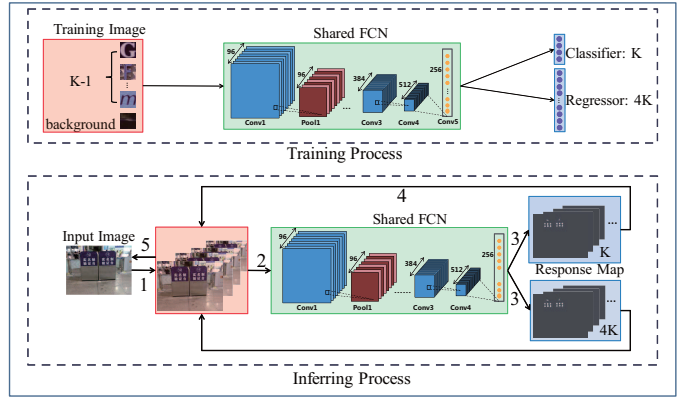}
   \caption{CPN contains training and inference processes. Training process is illustrated in Section 4. Testing process which consists of five steps is introduced in Section 5.}
   \label{fig:2}
\end{figure}

\section{Optimization}
For selecting training samples, positive samples contain two types: (1) those samples cropped by truth bounding boxes; (2) those samples which are shifted from truth bounding boxes. To preserve aspect ratio and avoid transformation, we expand the shorter side of each bounding box.
Further, we divide positive samples into $K$ templates based on their aspect ratios. Let $A=\{a_1,a_2,...,a_K\}$ denote $K$ types of aspect ratios. Let $\mathbf{\mathcal{M}}=\{\mathbf{m}_1,\mathbf{m}_2,..,\mathbf{m}_K\}$ denote a set of $K$ templates. The $k$-th template can be computed as:
\begin{equation}
 \mathbf{m}_k=\left\{
\begin{aligned}
&(\frac{R_w(1-a_k)}{2},R_h), & a_k < 1 \\
&(R_w, \frac{R_h(1-1/a_k)}{2}), & a_k \geq 1
\end{aligned}
\right.
\end{equation}
where $\mathbf{m}_k$ denotes the $k$-th template, $k\in \{1,2,..,K\}$.
On the other hand, negative samples are randomly cropped samples whose Intersection-over-Union (IoU) with any positive sample is less than 0.1. Positive and negative samples are resized to fixed-size and then fed into CPN-ENG or CPN-CHS.

For determining labels, in classification task, $c_j=k$ if the $j$-th training sample has more than 0.85 IoU with a true bounding box which belongs to $k$-th template, otherwise $c_j=K$.
For regression task, the labels of $i$-th sample are denoted as $t_x^{(i)}$, $t_y^{(i)}$, $t_w^{(i)}$ and $t_h^{(i)}$, which follows \cite{girshick2014rich}. They are computed as
\begin{equation}
t_x^{(i)} = (G_x^{(i)} - P_x^{(i)})/P_w^{(i)},
\end{equation}
\begin{equation}
t_y^{(i)} = (G_y^{(i)}-P_y^{(i)})/P_h^{(i)},
\end{equation}
\begin{equation}
t_w^{(i)}=log(G_w^{(i)}/P_w^{(i)}),
\end{equation}
\begin{equation}
t_h^{(i)}=log(G_h^{(i)}/P_h^{(i)}),
\end{equation}
where $\mathbf{P}_{x,y}^{(i)}=(P_x^{(i)}, P_y^{(i)})$ is the location of $i$-th training bounding box, while $\mathbf{P}_{w,h}^{(i)}=(P_w^{(i)}, P_h^{(i)})$ is the size of $i$-th training bounding box. Likewise, $\mathbf{G}_{x,y}^{(i)}=(G_x^{(i)}, G_y^{(i)})$ and $\mathbf{G}_{w,h}^{(i)}=(G_w^{(i)}, G_h^{(i)})$ are the location and size for $i$-th ground truth bounding box respectively.

We adopt softmax regression as the output layer of classification task, while cross-entropy is employed as the criterion for constructing loss $J(\mathbf{W}_{cls})$. On the other hand, we adopt a standard mean square error (MSE) as the loss function $J(\mathbf{W}_{reg})$ for location regression task following \cite{girshick2014rich}.
The loss function $J(\mathbf{W}_{cls})$ is written as
\begin{equation}
J(\mathbf{W}_{cls})=-\frac{1}{N}\sum_{i=1}^{N} \sum_{k=1}^{K}1\{{y^{(i)}=k}\}log \frac{e^{w_k^T x^{(i)}}} {\sum_{k=1}^K e^{w_k^T x^{(i)}} },
\end{equation}
where $K$ is the number of templates together with background, and $N$ is the number of training samples. The loss function $J(\mathbf{W}_{reg})$ is written as:
\begin{equation}
J(\mathbf{W}_{reg})=\frac{1}{N}\sum_{i=1}^{N}\sum_{k=1}^{K}(t^{(i)} - \mathbf{W}^T\phi_i)^2+\lambda \|\mathbf{W}\|^2,
\end{equation}
where $\phi_i$ denotes the feature of last convolutional layer.
The overall loss is comprised of classification loss and regression loss, which is formulated as:
\begin{equation}
J(\mathbf{W})=\alpha J(\mathbf{W}_{cls}) + (1 - \alpha)J(\mathbf{W}_{reg}),
\end{equation}
where $\alpha$ balances $J(\mathbf{W}_{cls})$ and $J(\mathbf{W}_{reg})$.

Our framework can be trained end-to-end by back-propagation and stochastic gradient descent (SGD), which is given in Algorithm 1.
\begin{algorithm}
\caption{Optimization method in training process}
\label{alg1}
\begin{algorithmic}[1]
\REQUIRE ~~\\
Iteration number $t=0$. Image patch $p_i$ and its corresponding labels $\{c_i, t_x^{(i)}, t_y^{(i)}, t_w^{(i)}, t_h^{(i)}\}$, $c_i\in \{1,...,K\}$.
\ENSURE ~~\\
Network parameters $\mathbf{W}$.
\REPEAT
    \STATE $t$ $\leftarrow$ $t+1$; \\
    \STATE Randomly select a subset of samples from training set.
    \FOR {each training sample}
    \STATE Do forward propagation to get $\phi_i = f(\mathbf{W}, p_i)$
    \ENDFOR
    \STATE¡¡$\Delta \mathbf{W}$ = 0
    \FOR {each training sample $\{p_i;c_i, x_i, y_i, w_i, h_i\}$}       
    \STATE Calculate partial derivative with respect to the output: $\frac{\partial J}{\partial \phi_i}$
    \STATE Run backward propagation to obtain the gradient with respect to the network parameters: $\Delta \mathbf{W}_i$
    \STATE Accumulate gradients: $\Delta \mathbf{W}: = \Delta \mathbf{W}+\Delta \mathbf{W}_i$
    \ENDFOR
    \STATE $\mathbf{W}^t:=\mathbf{W}^{t-1}-\epsilon_t \Delta \mathbf{W}$
\UNTIL{converge.}
\end{algorithmic}
\end{algorithm}

\section{Inference}
Our inference process containing five steps is presented in Fig. 2.

\noindent\textbf{Step 1}: Each testing image is resized to a series of images since the network is learned by fixed-size samples.
Multi-scale technique will ensure that the characters are activated in one of these scales.

\noindent\textbf{Step 2}: Multiple resized images are fed into CPN.

\noindent\textbf{Step 3}: Response maps for scores and locations are obtained.

\noindent\textbf{Step 4}:
(1) The coarse proposals can be obtained by using a linear mapping which is given as:
\begin{equation}
\mathbf{P}_{x,y}^{(i)}=s\times \mathbf{p}_{x,y}^{(i)} + \frac{\mathbf{R}-1}{2}
\end{equation}
where $s$ denotes the stride of two adjacent receptive fields. $\mathbf{R}=(29,29)$ for FCN-ENG or $\mathbf{R}=(43,43)$ for FCN-CHS. $\mathbf{p}_{x,y}^{(i)}$ is the coordinate of high-confidence unit in response map and $\mathbf{P}_{x,y}^{(i)}$ is the coordinate of the $i$-th proposal in scaled image. The size of corresponding patch is $\mathbf{P}_{w,h}^{(i)}=\mathbf{m}_k$.
(2) A refine process like \cite{girshick2014rich} is performed.
(3) Redundant proposals are filtered by Non-Maximal Suppression (NMS).

\noindent\textbf{Step 5}: Convert $\mathbf{P}_{x,y}^{(i)}$ from resized images into original image.
\section{Experiments}
\label{sec:typestyle}
\subsection{Experimental Settings}
\textbf{Datasets}: The ICDAR 2013 dataset \cite{karatzas2013icdar} includes 229 training images and 255 testing ones, while SVT dataset \cite{wang2010word} contains 100 training images and 249 testing images. We annotated the character bounding boxes in SVT dataset since SVT dataset has no character-level annotations. In addition to ICDAR 2013 and SVT dataset, we also perform evaluation on our Chinese dataset which is referred as Chinese2k.
We will release Chinese2k soon. The number of training characters for ICDAR 2013, SVT and Chinese2k dataset is 4619, 1638, 23768; while the number of testing character for ICDAR 2013, SVT and Chinese2k dataset is 5400, 4038, 4626.

\noindent\textbf{Evaluation criterion}: We use recall rate to evaluate different methods and it is formulated by:
\begin{equation}
recall=\frac{{\sum_{i=1}^M} \sum_{j=1}^{|T_i|} d(S_i, T_i^{j})}{\sum_{i=1}^M |T_i|}
\end{equation}
where $S_i$ denotes a set of proposals for $i$-th image, $T_i^j$ represents the $j$-th truth bounding box in $i$-th image. $|T_i|$ is the number of the truth bounding boxes in the $i$-th image. $d(S_i, T_i^{j})=1$ if at least one proposal in $S_i$ have more than a predefined IoU overlap with $T_i^j$, and 0 otherwise.

\noindent\textbf{Implementation Details}: In our experiments some public code of different methods were used with the following setup. For selective search (SS) \cite{uijlings2013selective} we use a fast mode and the minimum width of bounding boxes is set as 10 pixels. For Edge Boxes (EB) \cite{zitnick2014edge} we use the default parameters: step size of the sliding window is 0.65, and NMS threshold is 0.75; but we change the maximal number of bounding boxes to 5000. For MCG \cite{arbelaez2014multiscale} we use a fast mode. We can rank proposals in SS, EB and MCG by scores while MSER \cite{vedaldi2010vlfeat} can not. Hence we generate different number of MSER proposals through altering the delta parameter. We implement our learning algorithm based on Caffe framework \cite{jia2014caffe}. The min-batch size is 100 and the initial learning rate is 0.001. We use 18, 17 and 18 scales for ICDAR 2013, SVT and Chinese2k dataset respectively.
\begin{figure}[tb]
\includegraphics[width=0.5\textwidth]{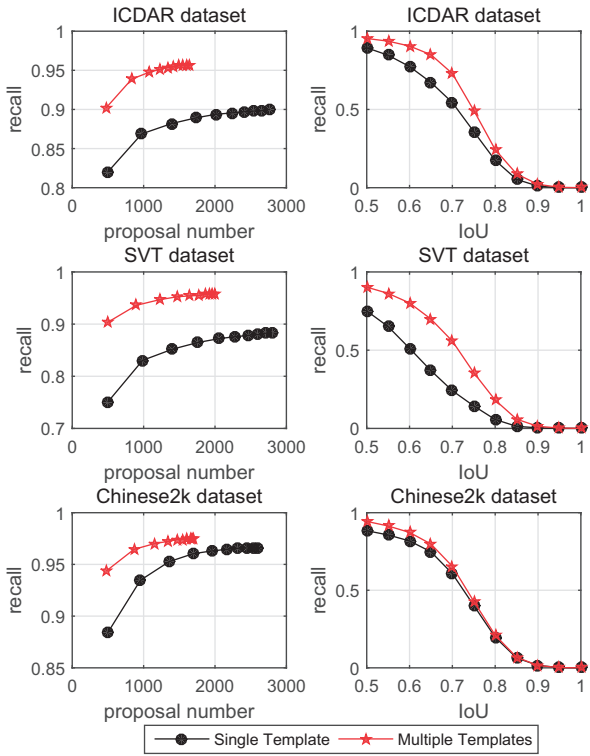}
   \caption{(a)Left column: recall rate vs. proposal number with single template and multiple template technique on three datasets (IoU=0.5). (b)Right column: Recall rate vs. IoU with single template and multiple template technique on three datasets (\#proposal=500).}
   \label{fig:4}
\end{figure}
\begin{figure}[tb]
\includegraphics[width=0.5\textwidth]{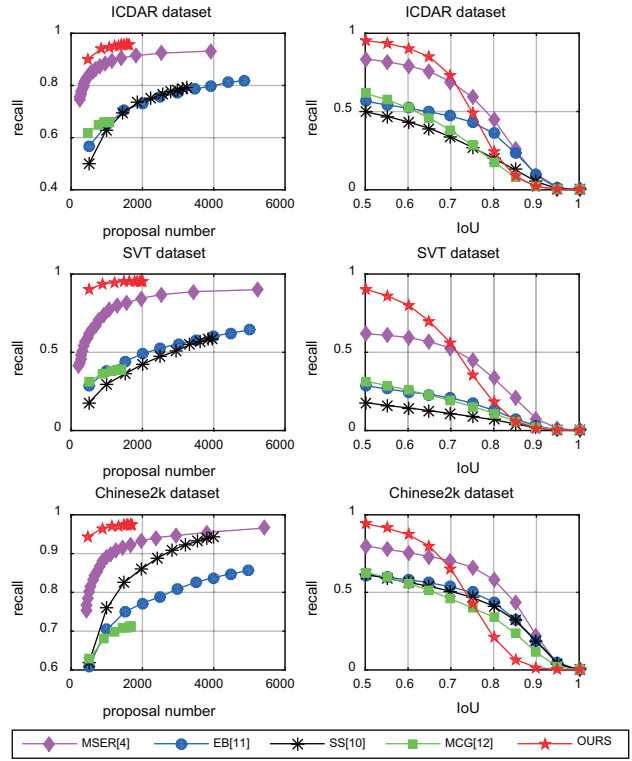}
   \caption{(a)Left column: recall rate vs. proposal number with our method and other state-of-the-arts on three datasets (IoU=0.5). (b)Right column: recall rate vs. IoU with our method and other state-of-the-arts on three datasets (\#proposal=500).}
   \label{fig:5}
\end{figure}
\subsection{Evaluation of Single Template and Multiple Templates}
Single template technique means that CPN outputs one template and background, namely $K=2$; while multiple template method used in our experiments means that CPN outputs $K-1$ templates and background, namely $K=4$. The comparison is evaluated on ICDAR 2013, SVT and Chinese2k dataset. We set hyper parameter $K$ as 4 (denote square-, thin-, flat-shape character and non-character) in these three datasets. Figure 3(a) indicates that the recall rate of multi-template technique is significantly higher than that of single template method, and Figure 3(b) shows that the IoU of multi-template technique is higher than that of single template when fixing recall rate. This is beneficial for the subsequent text location.
\subsection{Comparison with State of The Art}
It can be seen in Figure 4(a) our method outperforms all the evaluated algorithms in terms of recall rate on ICDAR 2013, SVT and Chinese2k dataset.
Moreover, our method achieves a recall rate of higher than $93\%$ even just using several hundreds of proposals, which will facilitate the succeeding procedures. It can be explained that our detector is a weak classifier for character and non-character classification while EB, SS and MCG methods are not designed for character proposals, hence our method surpass them under the case of less proposal number by a large margin. Further, we evaluate the performance of different proposal generators on a wide range of IoU. Figure 4(b) tells us that our method performs better than other methods in an IoU range of $[0.5, 0.7]$, while MSER is more robust in a more strict IoU range (e.g. $[0.7, 0.9]$).
%
%
\section{Conclusions}
\label{sec:majhead}
In this paper, we have introduced a novel method called character proposal network (CPN) based on fully convolutional network for locating character proposals in the wild. Moreover, a multi-template strategy integrating the different aspect ratios of scene characters is used. Experiments on three datasets suggest that our method convincingly outperforms other state-of-the-art methods, especially in the case of less proposal numbers and an IoU range of [0.5,0.7].
There are several directions in which we intend to extend this work in the future. The first is improve our model in a more strict IoU range. The second is to evaluate the effect for a text detection system.

\bibliographystyle{IEEEbib}
\bibliography{refs}

\end{document}